\documentclass[10pt,twocolumn,letterpaper]{article}

\usepackage[normalem]{ulem}  

\usepackage{cvpr} 
\usepackage{amsmath}
\usepackage{amssymb}
\usepackage{xcolor}  

\usepackage{pifont}
\newcommand{\red}[1]{{\color{red}#1}}

\definecolor{cvprblue}{rgb}{0.21,0.49,0.74}
\usepackage[pagebackref,breaklinks,colorlinks,allcolors=cvprblue]{hyperref}
\def\paperID{*****} 
\def\confName{CVPR}
\def\confYear{2025}

\title{EgoVLM: Policy Optimization for Egocentric Video Understanding}

\author{%
  Ashwin Vinod\thanks{All authors contributed equally.}\protect\footnotemark[1] ,\ 
  Shrey Pandit\protect\footnotemark[1] ,\ 
  Aditya Vavre\protect\footnotemark[1] ,\ 
  Linshen Liu\protect\footnotemark[1]\\
  The University of Texas at Austin\\
  \texttt{(ashwinv, shreypandit, aditya.vavre, linshen\_liu)@utexas.edu}
}

\begin{document}
\maketitle
\begin{abstract}
Emerging embodied AI applications, such as wearable cameras and autonomous agents, have underscored the need for robust reasoning from first person video streams. We introduce EgoVLM, a vision-language model specifically designed to integrate visual comprehension and spatial-temporal reasoning within egocentric video contexts. EgoVLM is fine-tuned via Group Relative Policy Optimization (GRPO), a reinforcement learning method adapted to align model outputs with human-like reasoning steps. Following DeepSeek R1-Zero’s approach, we directly tune using RL without any supervised fine-tuning phase on chain-of-thought (CoT) data.  We evaluate EgoVLM on egocentric video question answering benchmarks and show that domain-specific training substantially improves performance over general-purpose VLMs. Our EgoVLM-3B, trained exclusively on non-CoT egocentric data, outperforms the base Qwen2.5-VL 3B and 7B models by \textbf{+14.33} and \textbf{+13.87} accuracy points on the EgoSchema benchmark, respectively. By explicitly generating reasoning traces, EgoVLM enhances interpretability, making it well-suited for downstream applications. Furthermore, we introduce a novel keyframe-based reward that incorporates salient frame selection to guide reinforcement learning optimization. This reward formulation opens a promising avenue for future exploration in temporally grounded egocentric reasoning. Our Code can be found \href{https://github.com/adityavavre/VidEgoVLM}{here}.
\end{abstract}    
\section{Introduction}
\label{sec:intro}

Recent advancements in Vision-Language Models (VLMs) have primarily focused on third-person perspective images and videos, overlooking the unique challenges and opportunities presented in egocentric focused tasks. However, the rise of embodied AI technologies such as wearables, autonomous agents, and augmented reality (AR) systems, necessitates a shift toward models that can comprehend and reason from a first-person viewpoint. With the growing use of devices that offer an egocentric perspective, there is a significant demand for purpose-specific VLMs designed for reasoning in such contexts. Recent work on egocentric video understanding ~\cite{cheng2024videgothink, egotextvqa} underscores the limitations of current VLMs in handling first-person perspective data. This gap calls for the development of a robust, reasoning-oriented VLM tailored for egocentric videos—one that can enable AI systems to perceive, interpret, and respond to the world in a manner aligned with human perspective. Such a model would offer enhanced understanding of user intent, improved temporal and contextual awareness, and greater alignment with real-world human experiences.

In this work we seek to build a VLM endowed with reasoning capabilities, which is capable of harmoniously blending visual comprehension and spatial reasoning from an egocentric viewpoint within the video domain. Recent work in large language models (LLMs) has demonstrated the effectiveness of Group Relative Policy Optimization (GRPO) ~\cite{deepseek-math} in enhancing reasoning capabilities, particularly in mathematical and coding tasks ~\cite{deepseek-r1}. Adapting this technique, our study applies the GRPO training objective to VLMs to bridge existing gaps between visual perception and sophisticated reasoning tasks. Previous studies, such as MedVLM-R1~\cite{medvlmr1}, have demonstrated significant success by applying GRPO to medical VLMs, achieving enhanced generalization and computational efficiency compared to larger models, while explicitly generating reasoning processes alongside final answers. Inspired by these advancements, we extend GRPO to the egocentric video domain, where reasoning plays a central role in understanding user intent, and context. Unlike conventional supervised fine-tuning (SFT) approaches that may bias the model toward shallow pattern recognition, our method encourages the model to explore diverse reasoning paths and learn more robust decision-making strategies. We demonstrate that GRPO enables our model, EgoVLM, to achieve superior performance on complex egocentric benchmarks.

\begin{figure*}
\centering
    \includegraphics[width=1.0\linewidth]{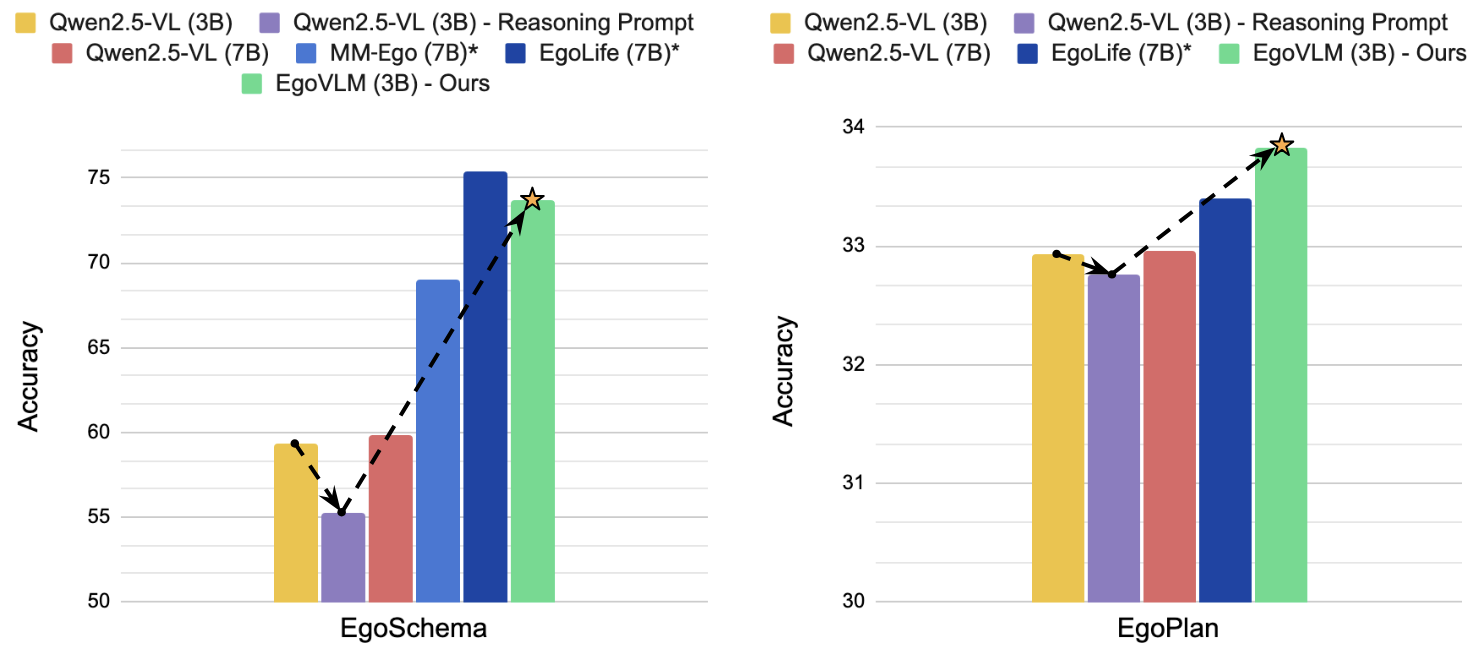}
    \caption{We significantly enhance the reasoning ability of the base Qwen2.5-VL (3B) model, surpassing even its 7B variant on two popular egocentric QA benchmarks, without the use of any CoT data. Additionally, our model outperforms the larger MM-Ego~\cite{mm-ego} model on the EgoSchema benchmark and exceeds the performance of EgoLife~\cite{egolife} on the EgoPlan benchmark. * indicates numbers borrowed from prior work. }\looseness=-1
    \label{fig:results_main}
\end{figure*}

\begin{figure*}
\centering
    \includegraphics[width=1.0\linewidth]{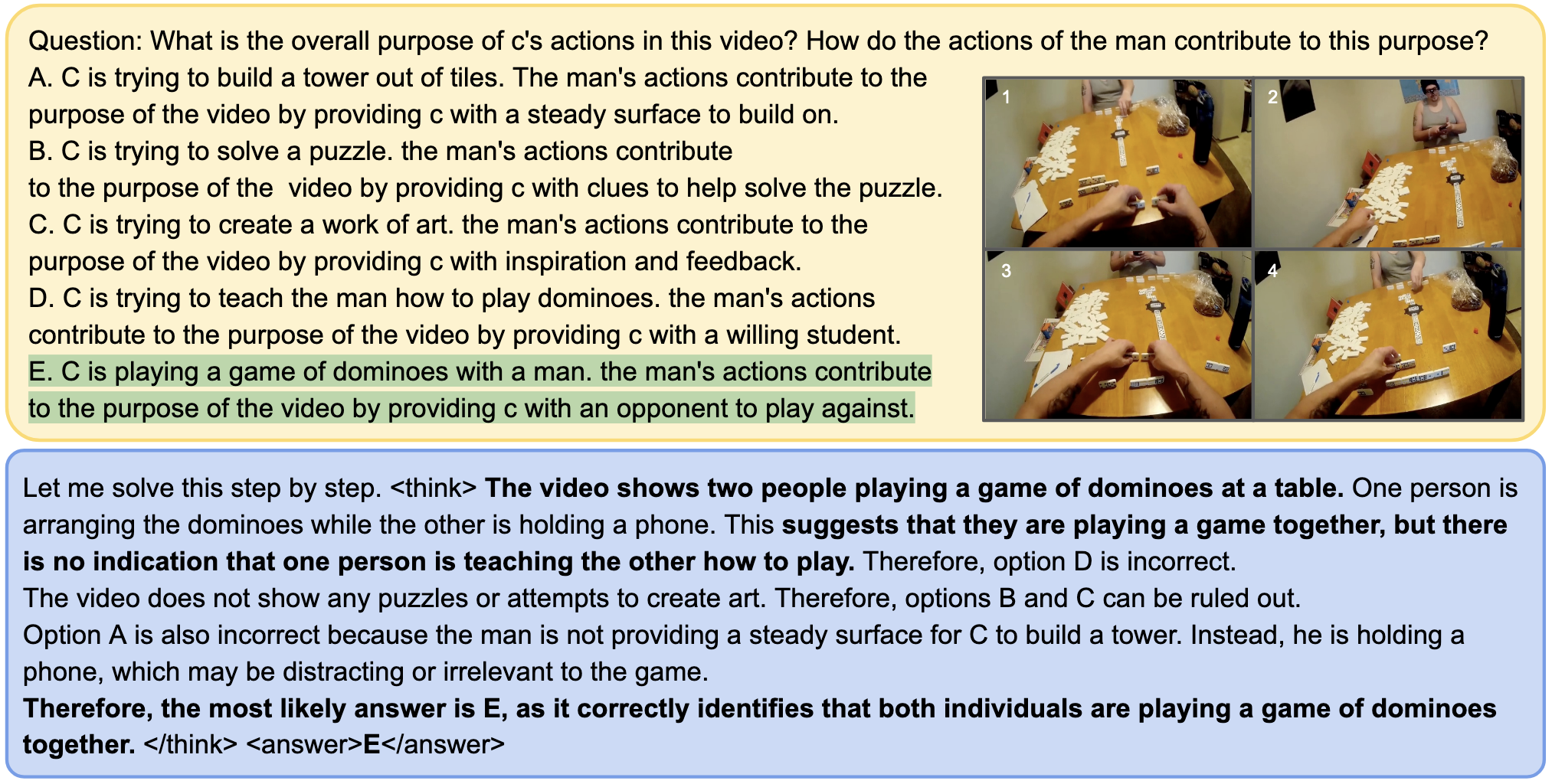}
    \caption{An example illustrating the reasoning capability of our EgoVLM model. The model successfully interprets the person's actions in the video and eliminates incorrect choices based on contextual understanding. The correct option is highlighted in green. }\looseness=-1
    \label{fig:example_main}
\end{figure*}

The key contributions of this paper are summarized as follows:
\begin{enumerate}
    \item We present EgoVLM (3B), a reasoning-focused vision-language model trained solely on \textbf{non-chain-of-thought (CoT)} egocentric data using the GRPO objective.
    \item We show that EgoVLM surpasses comparable and even larger-scale models (e.g., Qwen2.5-VL 3B and 7B) on egocentric question answering (QA) benchmarks. Our strong benchmark performance is highlighted in \autoref{fig:results_main}. 
    \item We also show that EgoVLM's strong reasoning capability generalizes well to non-egocentric videos. An example of the strong reasoning capability of EgoVLM is illustrated in \autoref{fig:example_main}.
    \item Through our ablation experiments we show that effective GRPO training with LoRA requires high model adaptation capacity and that GRPO training benefits from detailed prompts.
    \item We introduce a novel keyframe-based reward formulation integrated into the GRPO framework. While this approach does not lead to immediate performance gains on egocentric video understanding tasks, it offers a promising direction for future research.
\end{enumerate}
The rest of the paper is organized as follows: In \autoref{sec:related}, we review and present prior research effort in this area. In \autoref{sec:method}, we introduce the EgoVLM training objective. In \autoref{sec:setup}, we outline the experimental setup. In \autoref{sec:results}, we report our main experimental findings, including both quantitative metrics and qualitative visualizations. In \autoref{sec:keyframe} we discuss a novel keyframe reward formulation and its related experiments. In \autoref{sec:conclusion} we present the conclusion and discuss potential future work. 

\section{Related Work}
\label{sec:related}
\paragraph{RL Training Objectives}

Reinforcement learning (RL) has emerged as a powerful paradigm for training language models with reasoning capabilities, enabling models to align better with human preferences and downstream objectives. Among the most prominent RL techniques are Proximal Policy Optimization (PPO) \cite{schulman2017proximal}, Direct Preference Optimization (DPO) \cite{rafailov2023direct}, and Group Relative Policy Optimization (GRPO) \cite{shao2024deepseekmath}, each offering unique trade-offs in efficiency and alignment quality.

PPO optimizes policies using reward feedback and a separate value function, employed in models like InstructGPT\cite{ouyang2022training} and GPT-4\cite{achiam2023gpt}. DPO simplifies this by using pairwise preference comparisons without an explicit reward model, adopted by ChatGLM3-DPO\cite{glm2024chatglm} and WizardLM 2.0\cite{wizardlm2_2024} for efficient alignment signal capture.

Both approaches have limitations: PPO requires large auxiliary modules\cite{ahmadian2024back}, while DPO depends on carefully curated preference pairs\cite{li20252d}. GRPO offers a more efficient alternative by eliminating the value function and estimating advantages directly from group normalized reward outputs\cite{shao2024deepseekmath}. It has demonstrated strong performance in reasoning-centric applications like DeepSeekMath-RL\cite{shao2024deepseekmath} and excels in multi output reasoning tasks under resource constraints.

\paragraph{Reasoning VLMs} 
Recent reasoning-oriented VLMs are predominantly trained on heterogeneous datasets without viewpoint specificity, enabling broad visual reasoning but offering limited support for egocentric understanding. Notable examples include image-centric models such as Visual RFT~\citep{visualrft2025} and MedVLM-R1~\citep{medvlmr1}, which are trained on datasets like COCO~\citep{lin2014microsoft} and large-scale medical image corpora, respectively. While these models demonstrate strong performance on single-image reasoning tasks, they lack the temporal modeling necessary for video understanding. In contrast, our work will focus on egocentric video understanding, leveraging RL to enhance the reasoning abilities of VLMs in the video domain.
\paragraph{Egocentric VLMs}
Recent efforts in egocentric video understanding have focused on scaling data and introducing task-specific architectures. MM-Ego~\citep{mm-ego} presents a large-scale 7M-sample egocentric video QA corpus and introduces Memory Pointer Prompting, achieving state-of-the-art results on several benchmarks. EgoLife~\citep{egolife} contributes a 300-hour multi-participant dataset and proposes the EgoButler framework, which integrates multi-modal understanding with retrieval-augmented generation. However, both approaches lack general-purpose reasoning capabilities. In contrast, ST-Think~\citep{stthink} addresses spatio-temporal reasoning using chain-of-thought supervision and the GRPO objective, introducing the Ego-ST Bench and the ST-R1 model. While ST-Think shares similarities with our work in applying GRPO to egocentric data, its evaluation is limited to the benchmarks introduced in their study, without assessing performance on widely used existing egocentric QA datasets. Furthermore, their model relies on pretraining with chain-of-thought (CoT) data, whereas we demonstrate that strong reasoning performance can be achieved without such supervision.

\section{EgoVLM Method}
\label{sec:method}
In this section we describe the GRPO training objective that we use to train our EgoVLM model.

Group Relative Policy Optimization, or GRPO, is a reinforcement learning algorithm that improves a model's reasoning by using multiple sampled outputs for a given prompt and leveraging a reward function to fine-tune the model's behavior. To reduce the computational overhead associated with training an auxiliary value function model, as required by PPO~\cite{schulman2017proximal}, GRPO instead utilizes the mean reward from the policy model’s sampled responses as a baseline for advantage estimation. Specifically, for a given input question \(q\) we sample a group of completions \(\{o_i\}_{i=1}^G\) and compute their corresponding rewards \(\{r_i\}_{i=1}^G\) using the reward model. The advantage is then computed as:
\begin{equation}
\label{eqn1}
\hat{A}_{i,t}=\tilde{r}_i=\frac{r_i - \text{mean(\textbf{r})}}{\red{\text{std(\textbf{r})}}}
\end{equation}

The policy model is then optimized by maximizing the following KL objective:
\begin{equation}
\label{eq:grpo_aligned}
\begin{aligned}
J_{\text{GRPO}}(\theta)
&= \mathbb{E}_{q\sim P(Q)
   \,\{o_i\}_{i=1}^{G}\sim\pi_{\theta_{\mathrm{old}}}(O|q)}\Biggl[\\
&\frac{1}{G} \sum_{i=1}^{G} \red{{\frac{1}{\vert o_i\vert}}}\sum_{t=1}^{\vert o_i\vert}\Biggl\{
    \min\Biggl[\frac{\pi_\theta(o_{i,t}|q,o_{i<t})}{\pi_{\theta_{\mathrm{old}}}(o_{i,t}|q,o_{i<t})}\hat{A}_{i,t},\;\\
&\operatorname{clip}\Bigl(\frac{\pi_\theta(o_{i,t}|q,o_{i<t})}{\pi_{\theta_{\mathrm{old}}}(o_{i,t}|q,o_{i<t})}, 1-\epsilon,1+\epsilon\Bigr)\hat{A}_{i,t}\Biggr]\\
&\;-\;\beta\,\mathbb{D}_{\mathrm{KL}}\!\bigl(\pi_\theta\Vert\pi_{\mathrm{ref}}\bigr)
   \Biggr\}\Biggr]
\end{aligned}
\end{equation}
where $\pi_\theta$ and $\pi_{old}$ are the current and old policy, and $\epsilon$ and $\beta$ are hyper-parameters introduced in PPO.

\paragraph{Reward Modeling}
Consistent with the approach used in DeepSeek-R1, we avoid using a reward model and instead adopt a rule-based reward function that evaluates model outputs based on both structural compliance and answer correctness. Specifically, we use the following two reward functions: 
\begin{itemize}
    \item Format reward: If the model first provides a thinking process in \textless think\textgreater\ \textless/think\textgreater\ tags and then provides the final answer in \textless answer\textgreater\ \textless/answer\textgreater\ tags, it receives a format reward of 1.
    \item Accuracy reward: If the model provides a correct final answer within \textless answer\textgreater\ \textless/answer\textgreater\ tags, it receives an accuracy reward of 1. Note that for multiple-choice question answering (MCQA) and yes/no questions, correctness is evaluated based on exact string matching. In contrast, for open-ended question answering, we employ the ROUGE-L score\cite{rouge} as the accuracy reward in place of a binary signal.
\end{itemize}
Our final reward is determined as the sum of format and accuracy reward.
\paragraph{Dr. GRPO}
GRPO uses a group-wise baseline to estimate token-level advantages, but its original formulation divides the policy-loss term by both the length of each sampled response and the within-group standard deviation of returns. These normalizers introduce systematic biases: (i) the response-length bias causes correct but concise trajectories to receive disproportionately large weight while longer incorrect answers face less penalty, pushing the policy toward longer failures; (ii) the difficulty bias allows questions with small return variance to dominate mini-batch updates, distorting learning toward the extremes of the difficulty distribution.

Dr. GRPO (GRPO Done Right) \cite{drgrpo} simplifies GRPO by removing both normalization terms: it replaces per-response normalization with a fixed generation budget and drops per-question standard deviation scaling in the objective. The removed terms are \textcolor{red}{highlighted in red} in equations \ref{eqn1} and \ref{eq:grpo_aligned}. Motivated by the empirical success of Dr. GRPO reported in \cite{drgrpo}, we extend and evaluate a Dr. GRPO variant within our egocentric video reasoning framework.

%




\section{Experimental Setup}
\label{sec:setup}
We adopt \textbf{Qwen-2.5-VL-3B-Instruct} as our base model because of its strong vision-language capabilities, which we aim to further exploit through reinforcement learning. Training follows the GRPO objective and is implemented on top of the codebase from \cite{vlmr1}.  We conduct training with the following parameters: $\text{Training Steps} = 3200$, $\text{Group Size/\#Generations} = 6$, KL coefficient $\beta = 0.04$, Temperature $= 0.9$, and LoRA rank $= 1024$. Throughout training, we keep the vision encoder frozen. Due to compute constraints , We limit the number of generations during training to 6. We use a detailed prompt during training as shown in \autoref{tab:prompts}. It is important to note that GRPO training is typically preceded by a supervised fine-tuning (SFT) phase on long chain-of-thought (CoT) data, which helps the model learn to generate explicit reasoning traces. However, due to the absence of egocentric video CoT datasets, we omit this SFT step and adopt a setup similar to Deepseek’s R1-Zero~\cite{deepseek-r1}. We additionally train a variant with the Dr. GRPO~\cite{drgrpo} objective and a Supervised Fine-Tuning (SFT) baseline. We use LLaMA-Factory~\cite{llamafactory} for the SFT baseline while using the same LoRA rank and a frozen vision encoder. We use 8$\times$ GH200 nodes on the Vista compute cluster for training.

\paragraph{Training Dataset}
Our training corpus contains $\sim 27{,}000$ QA pairs drawn from $\sim 7{,}000$ video clips compiled from the EgoIT99K dataset~\cite{egolife}, yielding a QA dataset that is well-suited to the GRPO formulation. EgoIT99K is a 9 classic egocentric video datasets curated by \cite{egolife}. For more details on the composition of the dataset, please refer to \cite{egolife}. We restrict our dataset to contain only multi-choice (MCQA), Yes/No and a few open-ended questions. For each video, we sample 16 frames uniformly during training. We also restrict the input resolution to $64\times28\times28$ per frame during training due to compute constraints. 

\begin{table*}[ht!]
\centering
\small
\addtolength{\tabcolsep}{-1.5pt}
\begin{tabular}{lp{12cm}}
\toprule
\textbf{Prompt Template} & \textbf{Prompt} \\ \midrule \midrule
Simple & \textcolor{blue}{\{Question\}} First output a thinking process about the video and question within \textless think\textgreater\ \textless/think\textgreater\ tags and then output the final answer in \textless answer\textgreater\ \textless/answer\textgreater\ tags, e.g. \textless think\textgreater\ reasoning process here \textless/think\textgreater\ \textless answer\textgreater\ answer here \textless/answer\textgreater. \\
Detailed & \textcolor{blue}{\{Question\}} First analyze and think about the given video and question. Provide a detailed thinking process about the video within \textless think\textgreater\ \textless/think\textgreater\ tags and then output the final answer in \textless answer\textgreater\ \textless/answer\textgreater\ tags, e.g. \textless think\textgreater\ thinking process here \textless/think\textgreater\ \textless answer\textgreater\ answer here \textless/answer\textgreater. While thinking you must robustly verify your solution. It's encouraged to include self-reflection or verification in the thinking process. \\
Detailed w/ Keyframes & \textcolor{blue}{\{Question\}} Analyze and think about the given video and question. The given video contains 16/32 frames. First provide a detailed thinking process about the video, how it relates to the question, and which frames contain important information within \textless think\textgreater\ \textless/think\textgreater\ tags. Then pick \textcolor{blue}{\{k\}} important frames from the video needed to answer the given question and list their indices (1-based, comma-separated) within \textless frames\textgreater\ \textless/frames\textgreater\ tags. Finally provide the answer within \textless answer\textgreater\ \textless/answer\textgreater\ tags. For example, \textless think\textgreater\ thinking process here \textless/think\textgreater\ \textless frames\textgreater\ frame indices here \textless/frames\textgreater\ \textless answer\textgreater\ answer here \textless/answer\textgreater. While thinking you must robustly verify your solution. It's encouraged to include self-reflection or verification in the thinking process. \\
Detailed w/ Irrelevant frames & \textcolor{blue}{\{Question\}} Analyze and think about the given video and question. The given video contains 16/32 frames. First provide a detailed thinking process about the video, explain how it relates to the question, and locate frames in the video that are irrelevant to the question. Provide this thinking process within \textless think\textgreater\ \textless/think\textgreater\ tags. Then list the non-informative frames indices (1-based, comma-separated) within \textless frames\textgreater\ \textless/frames\textgreater\ tags. Finally provide the answer within \textless answer\textgreater\ \textless/answer\textgreater\ tags. For example, \textless think\textgreater\ thinking process here \textless/think\textgreater\ \textless frames\textgreater\ irrelevant frame indices here \textless/frames\textgreater\ \textless answer\textgreater\ answer here \textless/answer\textgreater. While thinking, you must filter out irrelevant visual information and robustly verify your solution. It is encouraged to include self-reflection or verification in the thinking process.
\\ \bottomrule
\end{tabular}
\caption{Prompt templates that are used during GRPO training. Each question is inserted into the template which serves as the model input both during training and inference. We use the detailed prompt template for EgoVLM. }
\label{tab:prompts}
\end{table*}


\paragraph{Evaluation Benchmarks}
We evaluate on three popular MCQA benchmarks and report accuracy: \textbf{EgoSchema}~\cite{mangalam2023egoschema} (test set), \textbf{EgoPlan}~\cite{chen2023egoplan} (validation set), and \textbf{MV-Bench}~\cite{li2024mvbenchcomprehensivemultimodalvideo}. \textbf{EgoPlan} contains 3,355 questions from real-world egocentric videos. The benchmark requires next-action inference from historical clips and a final frame, highlighting long-horizon reasoning in 419 distinct scenes. We use the validation set for our evaluation because the test set ground truth is not publicly available. For \textbf{EgoSchema}, we evaluate our model on the test set which consists of 5031 samples. While EgoSchema and EgoPlan are egocentric, \textbf{MV-Bench} contains exocentric videos; including it allows us to measure how well our egocentrically-trained reasoning model transfers to third-person scenarios. We increase the frame sampling rate to 32 frames per video and also increase the frame resolution to $128\times28\times28$ per frame during inference to boost performance.

\section{Results}
\label{sec:results}

\begin{figure*}[htp]
\centering
\includegraphics[width=.33\textwidth]{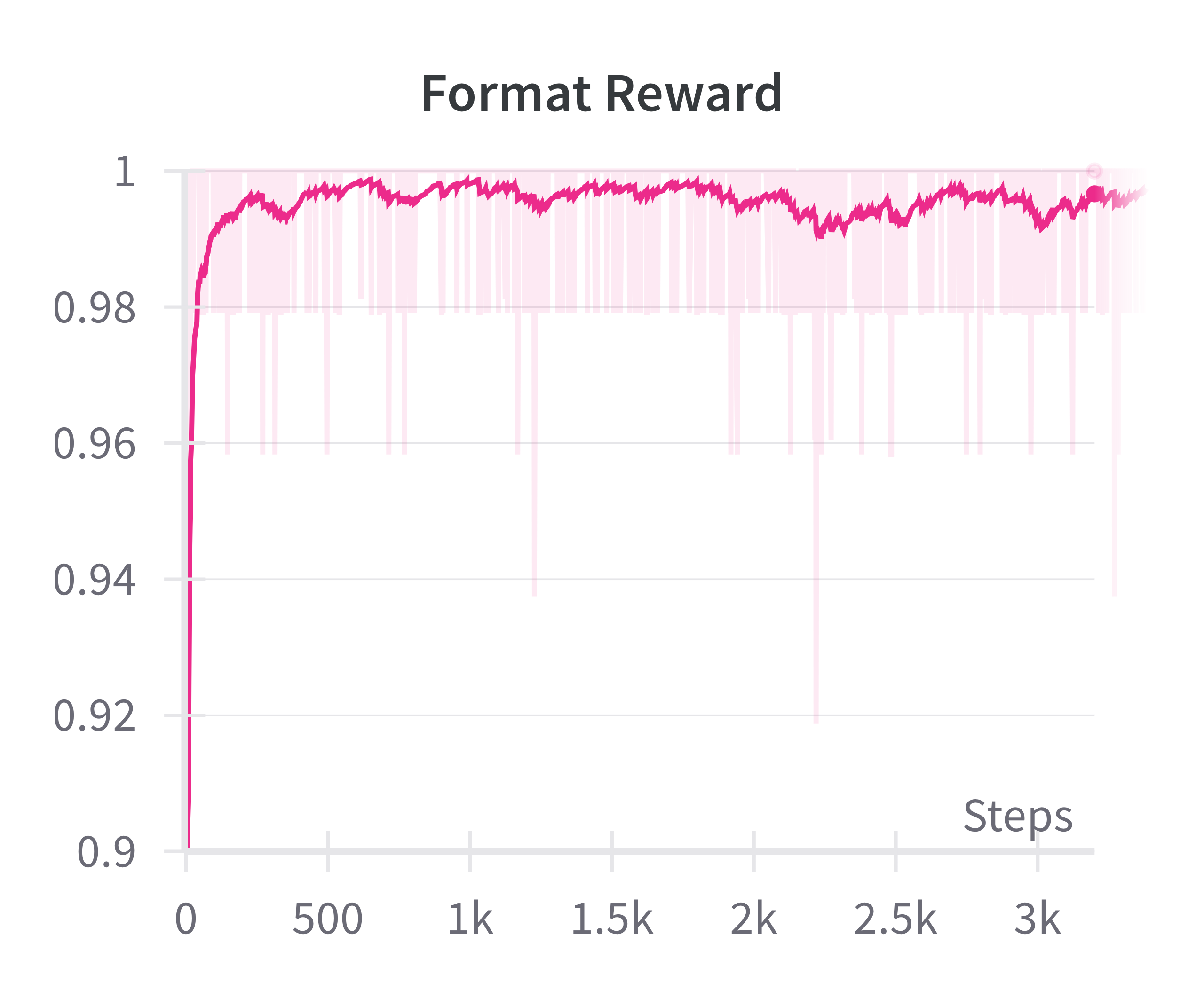}\hfill
\includegraphics[width=.33\textwidth]{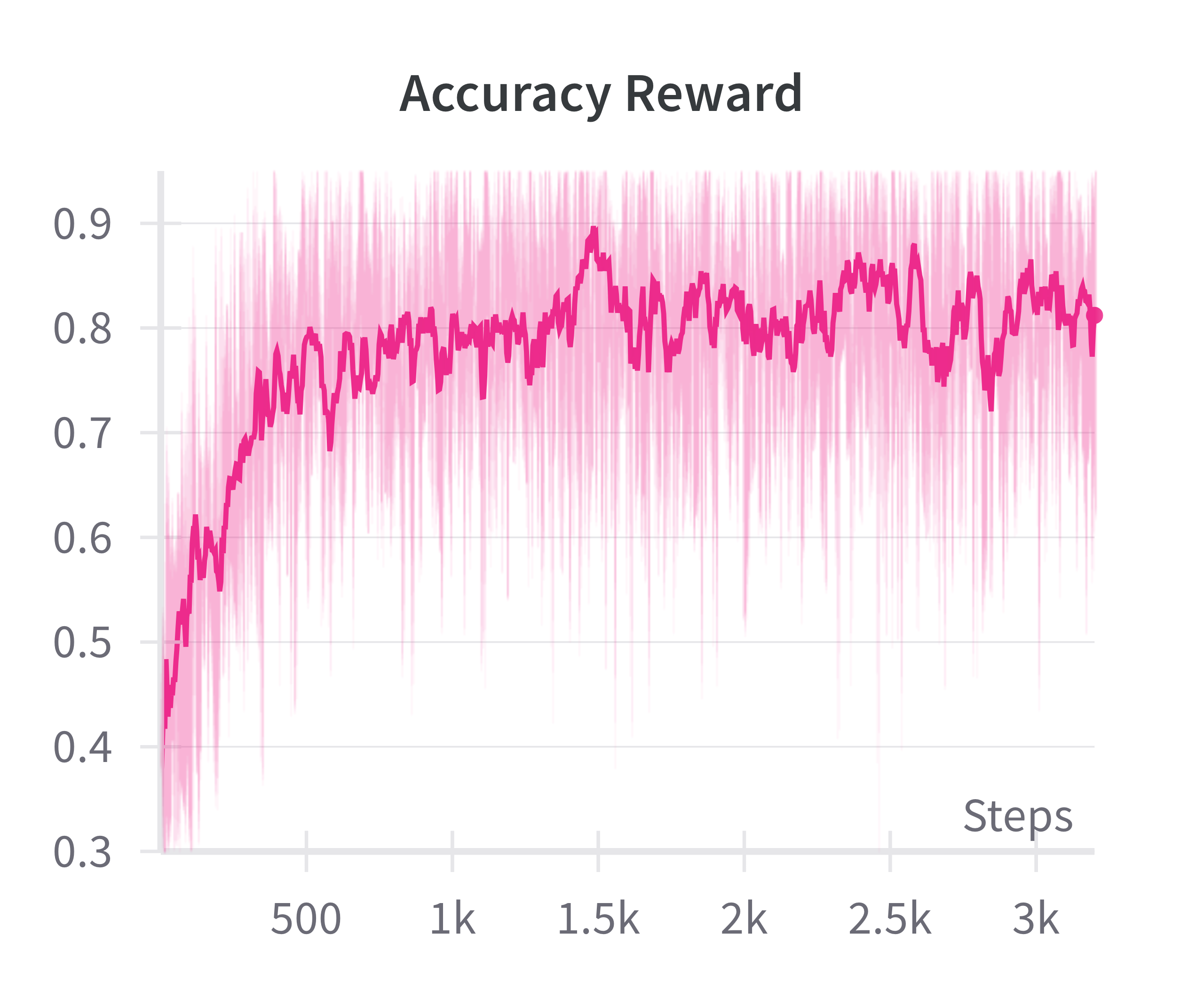}\hfill
\includegraphics[width=.33\textwidth]{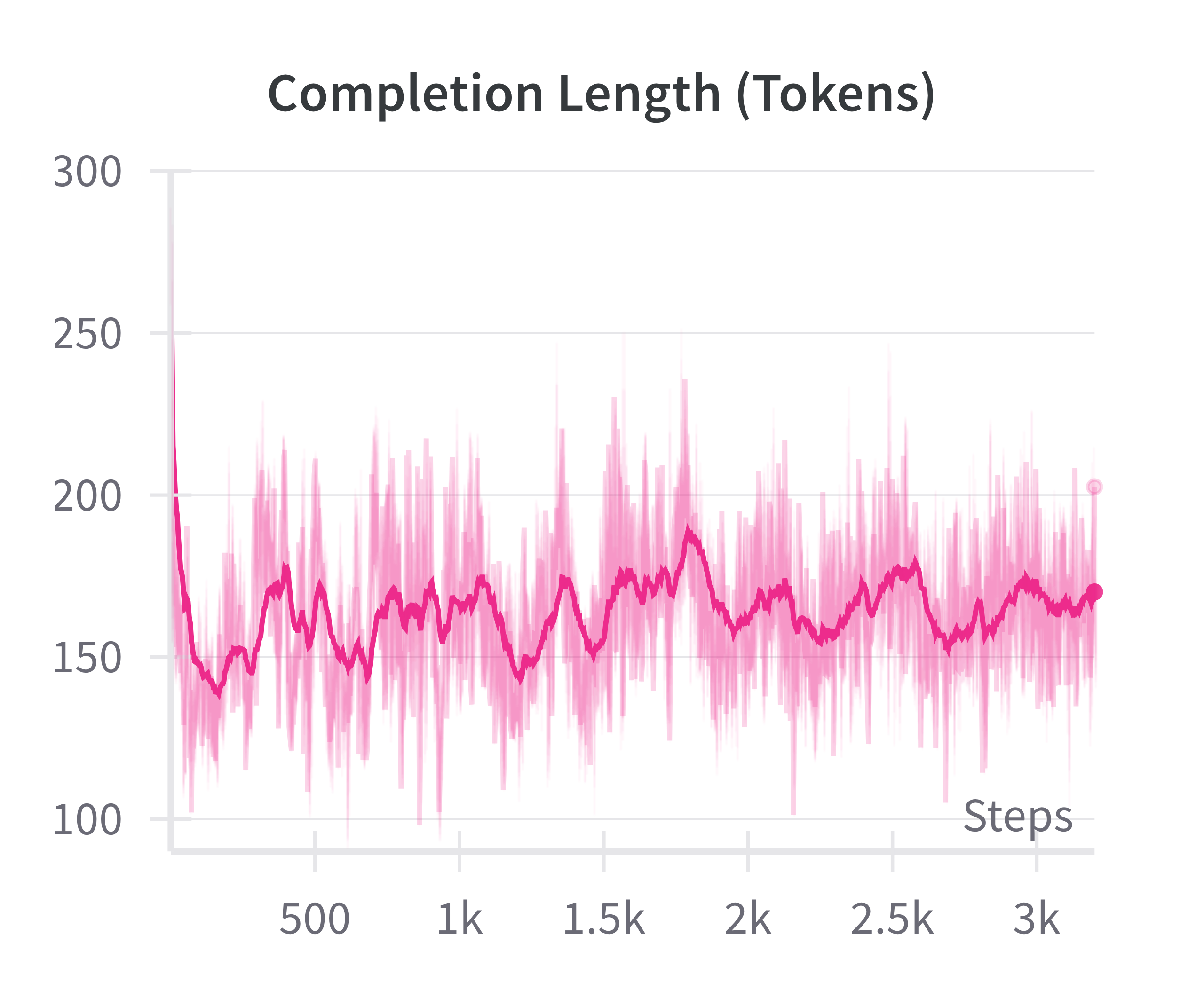}
\caption{Training visualization of EgoVLM with the GRPO training objective. Completion length refers to the length of generated output.}\looseness=-1
\label{fig:training_charts}
\end{figure*}

\begin{table*}[ht!]
\centering
\small
\addtolength{\tabcolsep}{-1.5pt}
\begin{tabular}{lccccc}
\toprule
\textbf{Model} & \textbf{\#Param} & \textbf{EgoSchema}$\uparrow$ & \textbf{EgoPlan}$\uparrow$ & \textbf{MVBench}$\uparrow$ & \textbf{Average}$\uparrow$ \\ \midrule \midrule
\textit{Proprietary} \\
GPT-4v*~\cite{gpt4v} & - & 56.6 & 38.0 & 43.5 & 46.0 \\
Gemini1.5-Pro*~\cite{gemini} & - & 72.2 & 32.8 & 60.5 & 55.2 \\
GPT-4o*~\cite{gpt4o} & - & 72.2 & 32.8 & 49.1 & 51.4 \\
\midrule
\textit{Open-source MLLMs} \\
Qwen2.5-VL~\cite{qwen2.5VL} & 3B &  59.4 &	32.9 & 66.9 & 53.1 \\ 
Qwen2.5-VL~\cite{qwen2.5VL} & 7B &  59.8 & 33.0 & 67.8 & 53.5\\
InternVideo2.5*~\cite{internvideo2.5} & 7B & 63.9 & - & 75.7 & - \\
LLaVa-Video*~\cite{llavavideo} & 7B & 53.7 & \underline{33.6} & 58.6 & 48.6 \\
\midrule
\textit{Ours} \\
SFT & 3B & \textbf{74.0} & 32.1	& 66.9 & 57.7 \\
Dr. GRPO & 3B & 68.4 & 33.0 & 64.8 & 55.4 \\
EgoVLM (GRPO) & 3B & \underline{73.7} & \textbf{33.8} & 66.5 & \textbf{58.0} \\ \bottomrule
\end{tabular}
\caption{Accuracy (in \%) on EgoSchema, EgoPlan and MVBench. This table compares our GRPO model with several state-of-the-art commercial and open-source models. Note that the reported performance on EgoPlan is on it's validation set. *indicates numbers borrowed from prior work.}
\label{tab:main_results}
\end{table*}

In this section, we present both quantitative and qualitative results from our experiments. 
\paragraph{Training Visualization} 
First, we illustrate the training dynamics of EgoVLM model in the plots of accuracy reward, completion length, and format reward as seen in \autoref{fig:training_charts}. The format reward quickly climbs to its upper bound, reflecting consistent adherence to the expected output structure throughout training. This is likely a result of using an instruction-tuned base model, which is already optimized to generate responses in the specified format early in training. The accuracy reward curve shows a clear upward trend indicating that the model progressively learns to generate more accurate answers. However, the completion length chart shows a clear absence of the commonly observed "Aha" moment during GRPO training. The "Aha" moment is an inflection point which is associated with the model learning to re-evaluate and refine its reasoning process, often accompanied by a sharp increase in completion length~\cite{zhou2025r1zerosahamomentvisual}. This is not observed even though we explicitly encourage the model to include self-reflection in its thinking process (see \autoref{tab:prompts}). Instead, the completion length remains relatively stable, without any such transition. This suggests that while the model learns to optimize for correctness and format, it does not develop a strong preference for brevity or efficiency in its outputs under the current training setup.

\paragraph{Benchmark Perfromance}
As shown in \autoref{tab:main_results}, we report the performance of EgoVLM, trained with GRPO, alongside a SFT baseline on the EgoSchema, EgoPlan, and MVBench benchmarks. We also compare GRPO with the more recent Dr. GRPO~\cite{drgrpo} variant. EgoVLM achieves state-of-the-art accuracy on the EgoSchema benchmark, outperforming all proprietary and open-source multimodal large language models (MLLMs), with only a slight performance gap compared to the SFT-trained variant. Notably, on the more challenging EgoPlan validation benchmark, which requires models to reason about a user's subsequent actions toward achieving a specified goal, EgoVLM surpasses all open-source baselines. We attribute this to GRPO’s enhanced ability to model complex, goal-directed reasoning. In contrast, SFT appears to degrade performance on EgoPlan, suggesting it may lack sufficient inductive bias for multi-step planning. Additionally, we observe that Dr. GRPO consistently underperforms relative to GRPO across all benchmarks. This performance gap is likely due to the small group size (6) used during training, which increases the variance in advantage estimates due to noisier baseline means in Dr. GRPO. Without GRPO’s stabilizing effect from standard deviation normalization, this heightened variance may lead to unstable policy updates and reduced overall performance.
EgoVLM also demonstrates strong generalization to non-egocentric video data, as shown by its competitive performance on the MVBench benchmark. However, since MVBench features tasks such as object counting, motion direction, and character sequencing which rely less on reasoning and more on perceptual precision, the model exhibits a slight decline in performance. 

\begin{figure*}
\centering
    \includegraphics[width=1.0\linewidth]{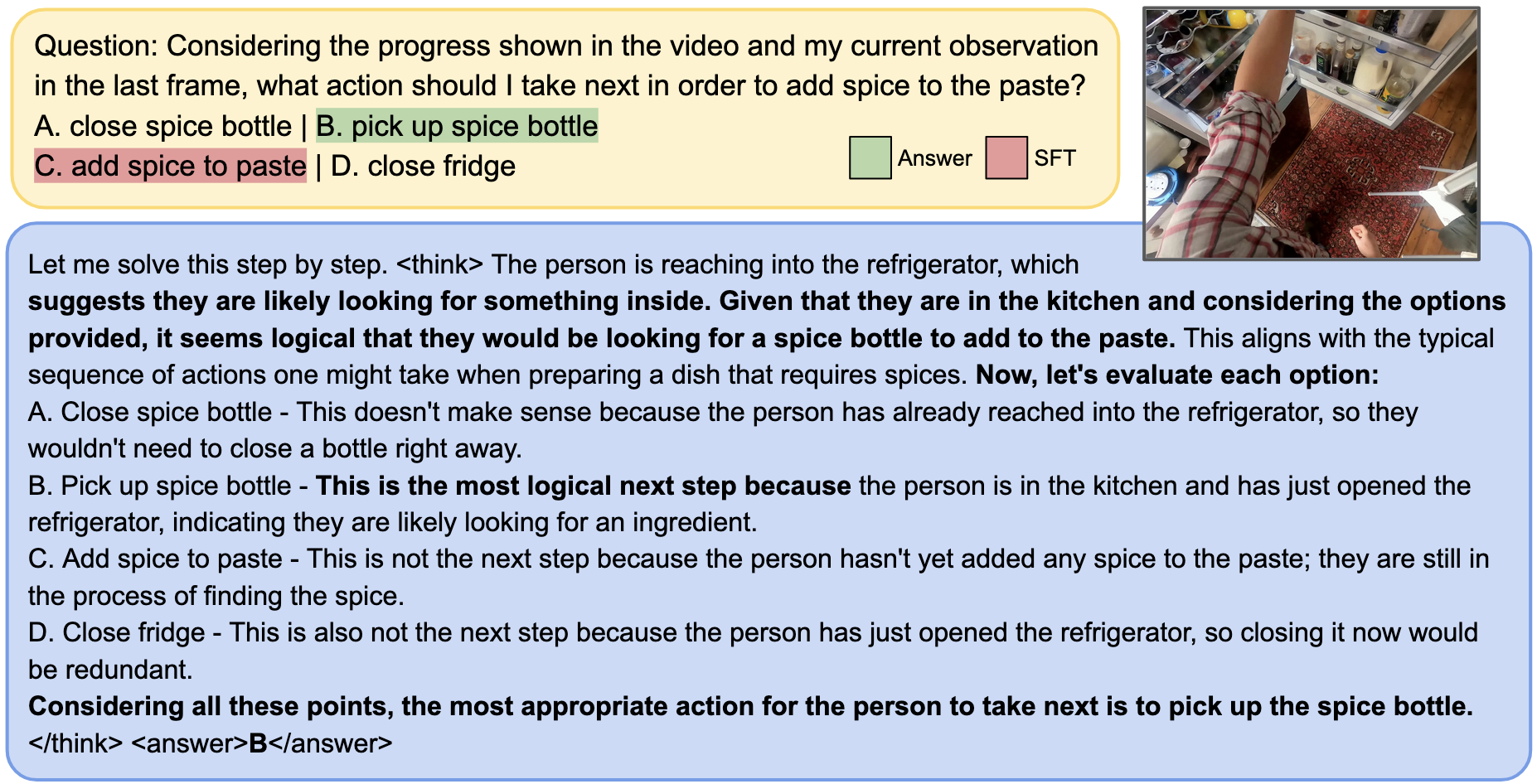}
    \caption{An example illustrating the reasoning capability of our EgoVLM model. EgoVLM is able to reason about a person's intent, correctly identifying that they must be looking for something in the fridge.}\looseness=-1
    \label{fig:example_qual_1}
\end{figure*}
\begin{figure*}
\centering
    \includegraphics[width=1.0\linewidth]{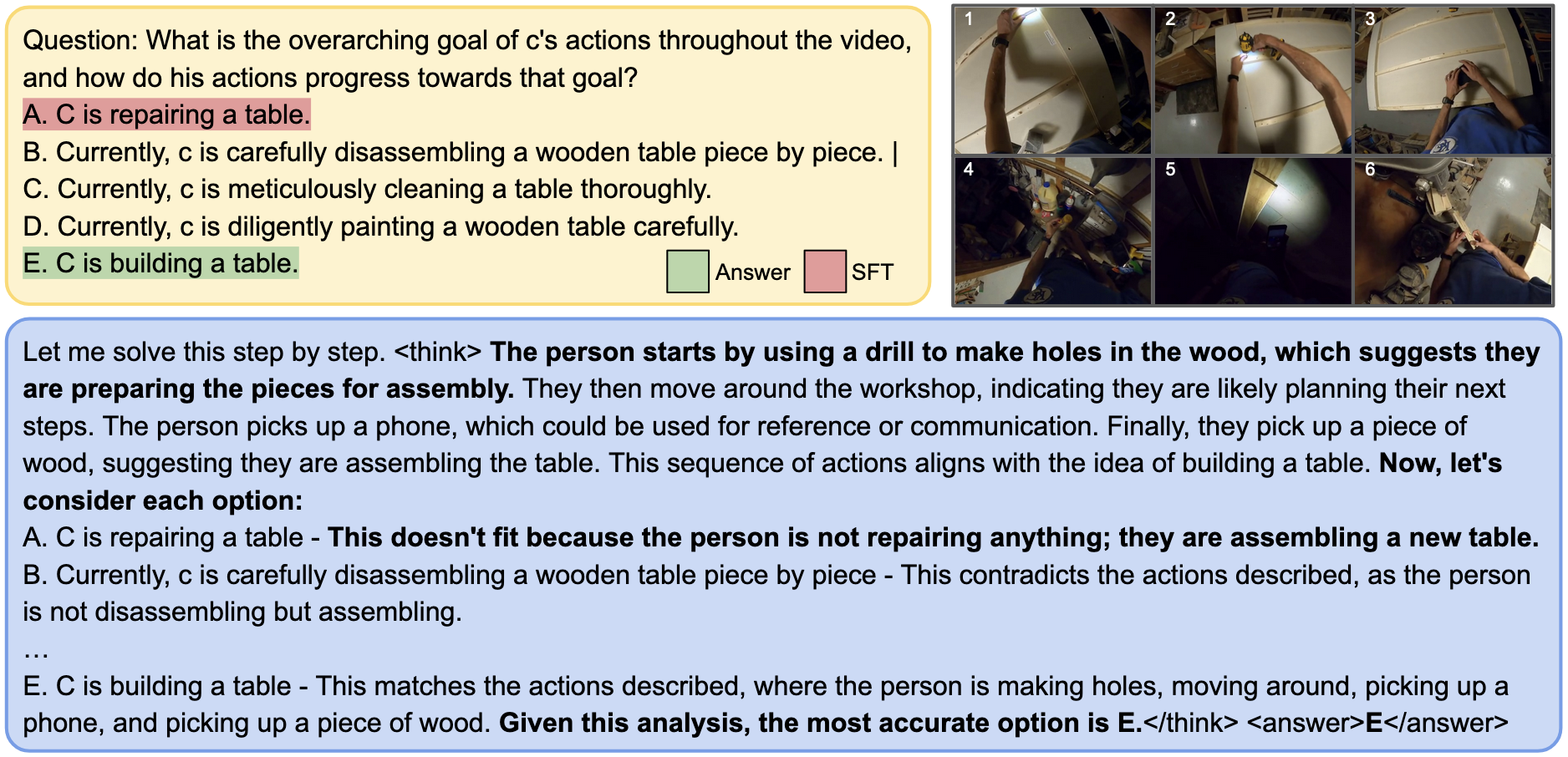}
    \caption{An example illustrating the reasoning capcbility of our EgoVLM model. EgoVLM reasons and disambiguates between two closely related actions.}\looseness=-1
    \label{fig:example_qual_2}
\end{figure*}
\begin{figure*}
\centering
    \includegraphics[width=1.0\linewidth]{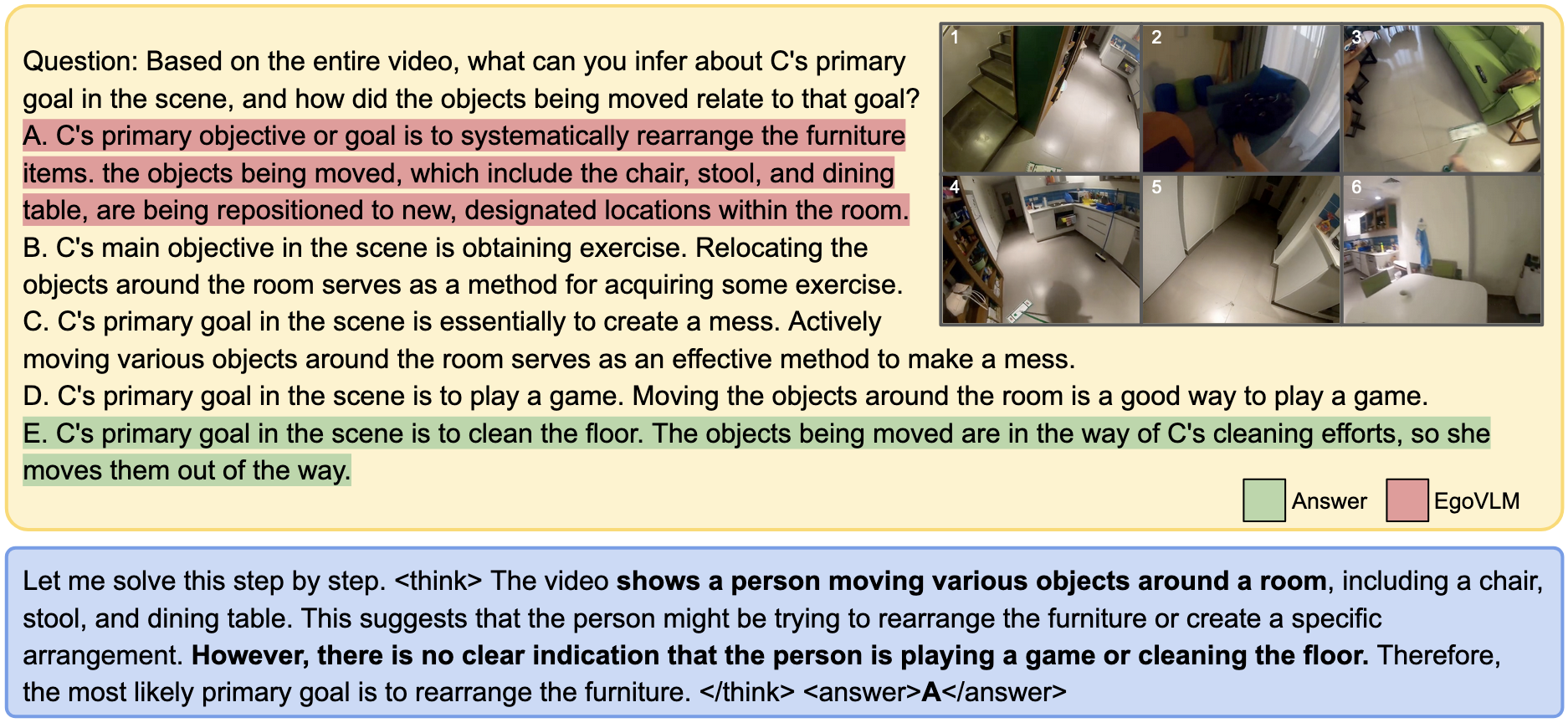}
    \caption{A failure case of our EgoVLM model. EgoVLM fails to identify a mop in the video. }\looseness=-1
    \label{fig:example_qual_3}
\end{figure*}
\begin{figure*}
\centering
    \includegraphics[width=1.0\linewidth]{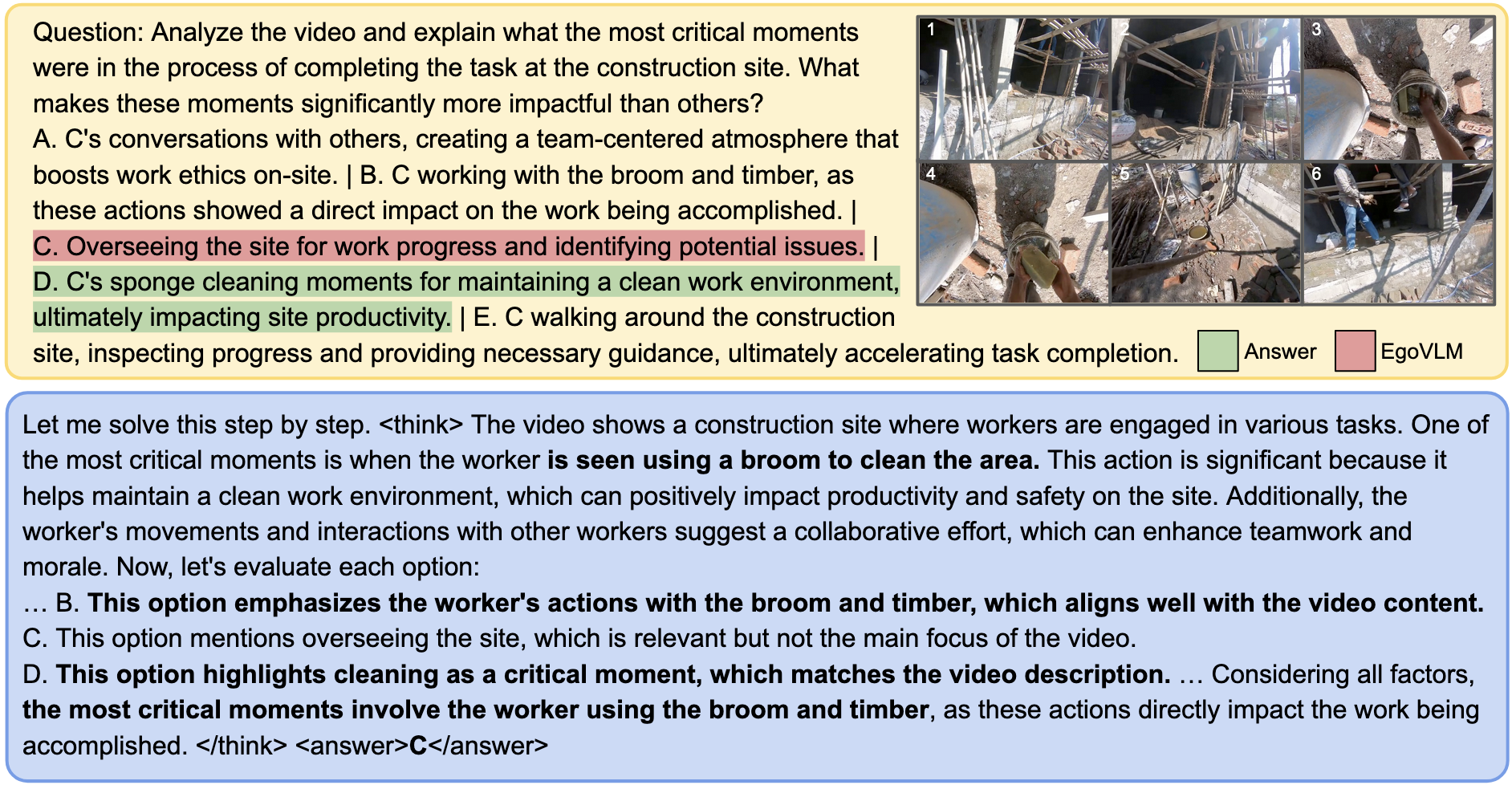}
    \caption{A failure case of our EgoVLM model. EgoVLM produces answers that are inconsistent with its reasoning traces.}\looseness=-1
    \label{fig:example_qual_4}
\end{figure*}

\paragraph{Qualitative Analysis}
We further illustrate the reasoning capabilities of our model through qualitative examples in \autoref{fig:example_qual_1} and \autoref{fig:example_qual_2}. In \autoref{fig:example_qual_1}, drawn from the EgoPlan benchmark, EgoVLM demonstrates a nuanced understanding of the user’s intent by correctly identifying the appropriate next action in a goal-oriented task. In contrast, the SFT-trained model fails to account for the underlying context and instead selects an option that prematurely reflects the user's final objective, rather than the required intermediate step. Similarly, \autoref{fig:example_qual_2}, from the EgoSchema benchmark, showcases EgoVLM’s ability to disambiguate between closely related actions. Specifically, it is able to distinguish between ‘building’ and ‘repairing’ by leveraging contextual reasoning cues. The SFT model, by comparison, misclassifies the scenario, highlighting its limited ability to perform fine-grained reasoning. 
Hence, in addition to its strong performance, EgoVLM offers greater explainability and trustworthiness compared to the baseline models. The GRPO training framework encourages the generation of diverse and interpretable reasoning paths, allowing users to better understand the model’s decision-making process which is an essential feature for deployment in real-world egocentric applications. 
\paragraph{Limitations} While EgoVLM demonstrates strong reasoning capabilities, we also observe certain limitations. As shown in \autoref{fig:example_qual_3}, the model fails to recognize a mop present in the input video. This failure may stem from factors such as occlusion, motion blur, or the relatively small size of the object, highlighting a current limitation in detecting fine-grained visual details. One potential direction for addressing this issue is to unfreeze and fine-tune the vision encoder during GRPO training. This could better adapt the visual backbone to egocentric scenarios, enabling it to capture more task-relevant details and improve end-to-end performance. Another example is provided in \autoref{fig:example_qual_4}, where the model's final prediction is not fully consistent with its reasoning trace. Moreover, the model occasionally hallucinates visual details, such as incorrectly recognizing the presence of a broom that is clearly absent from the video. Such discrepancies indicate that there remains room for improvement in aligning the model’s reasoning steps with its final outputs.

\subsection{Ablation Experiments}
This section presents some ablation experiments conducted to analyze the contribution of key design choices in our approach.
\begin{table}[ht!]
\centering
\small
\addtolength{\tabcolsep}{-4.5pt}
\begin{tabular}{lcccc}
\toprule
\textbf{Model} & \textbf{Reasoning} & \textbf{EgoSchema}$\uparrow$ & \textbf{EgoPlan}$\uparrow$ & \textbf{MVBench}$\uparrow$  \\ \midrule \midrule
Qwen2.5-VL & \ding{55} & 59.4 & 32.9 & 66.9 \\ 
EgoVLM & \ding{51} & \textbf{73.7} & \textbf{33.8} & 66.5 \\
EgoVLM & \ding{55} & 72.2 & 32.1 & \textbf{67.2} \\
\bottomrule
\end{tabular}
\caption{Results (Accuracy in \%) of different inference approaches on EgoVLM. Inference without a reasoning step improves performance on MVBench.}\looseness=-1
\label{tab:ablations_inference}
\end{table}

\begin{table}[ht!]
\centering
\small
\addtolength{\tabcolsep}{-5.pt}
\begin{tabular}{lccccc}
\toprule
\textbf{Model} & \textbf{EgoSchema}$\uparrow$ & \textbf{EgoPlan}$\uparrow$ & \textbf{MVBench}$\uparrow$ & \textbf{Avg.}$\uparrow$ \\ \midrule \midrule
Qwen2.5-VL &  59.4 & 32.9 & \textbf{66.9} & 53.1 \\ 
EgoVLM & \textbf{73.7} & 33.3 & 65.9 & \textbf{57.7} \\ 
LoRA $r=256$ & 66.6 & \textbf{33.6} & 64.3 & 54.8 \\
Simple Prompt & 70.2 & 33.2 & 64.8 & 56.0 \\ \bottomrule
\end{tabular}
\caption{Results (Accuracy in \%) of our ablation experiments. Note that these results denote performance after only 1000 training steps.}\looseness=-1
\label{tab:ablations}
\end{table}

\paragraph{Inference Setup} 
Recent work on reasoning LLMs~\cite{ma2025reasoningmodelseffectivethinking} has shown that bypassing the thinking process via simple prompting during the inference stage is surprisingly effective and computationally efficient. Motivated by this, we re-evaluated EgoVLM using a prompt that instructs the model to directly predict the answer without generating an intermediate reasoning trace. The results, presented in \autoref{tab:ablations_inference}, show a noticeable decline in performance on egocentric benchmarks, underscoring the critical role of reasoning in egocentric video understanding. Interestingly, this inference setup leads to improved results on MVBench, even surpassing the performance of the base model. This can be attributed to the nature of MVBench tasks, which primarily emphasize perceptual precision over complex reasoning.

\begin{figure}[h!]
\centering
\includegraphics[width=\linewidth]{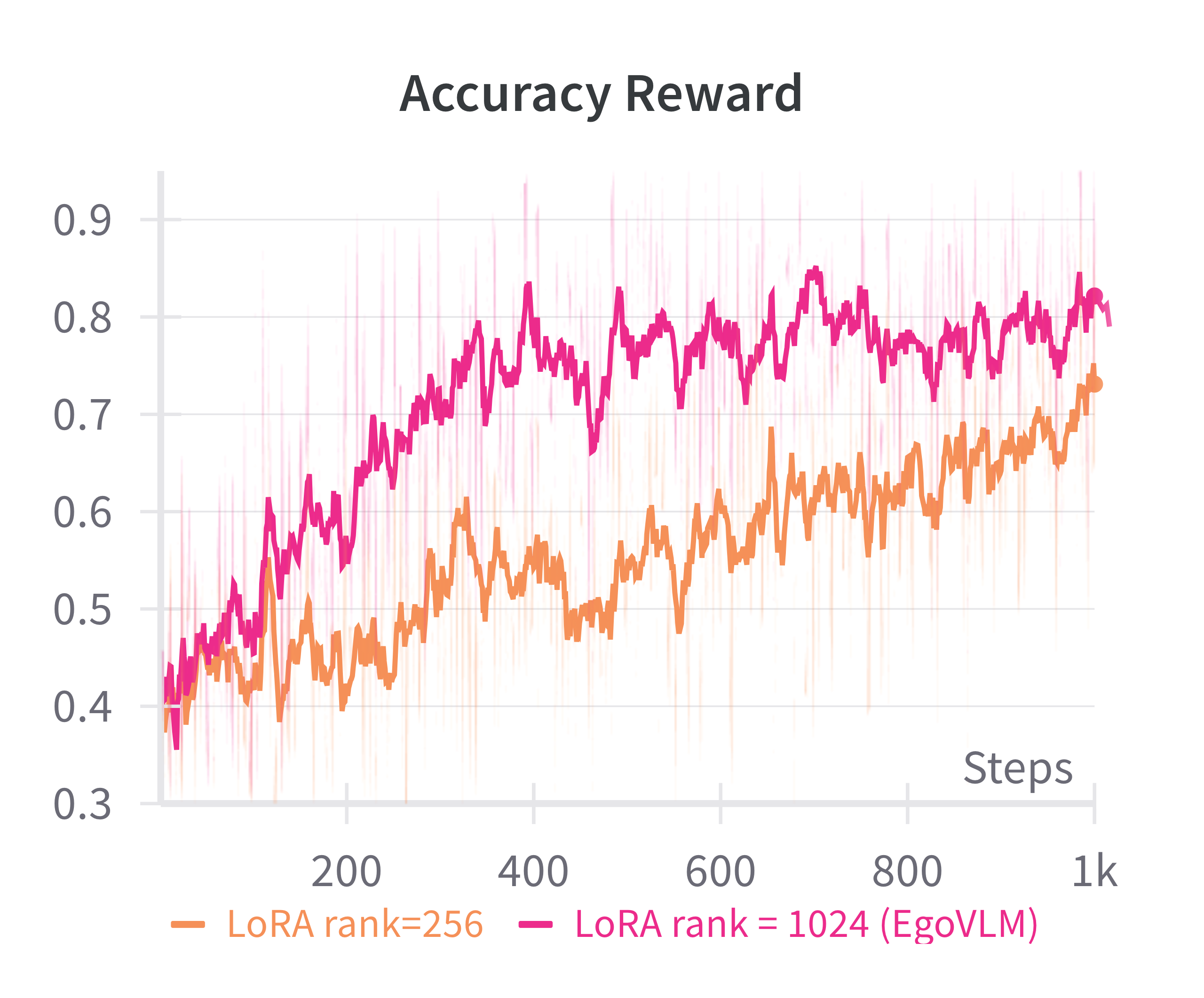}
\caption{Accuracy reward with different LoRA ranks in GRPO training.}\looseness=-1
\label{fig:lora_acc_reward}
\end{figure}
\paragraph{Others}
We further investigate the impact of model training capacity and prompt complexity by training EgoVLM using a reduced LoRA rank of 256 and a simplified prompt (refer to \autoref{tab:prompts}) within the GRPO framework. Due to computational constraints, these experiments were limited to 1,000 training steps. As shown in the accuracy reward curve in \autoref{fig:lora_acc_reward}, the configuration with the smaller LoRA rank exhibits significantly slower convergence compared to the original setup. We report the performance of these runs in \autoref{tab:ablations}. It is evident that the EgoVLM model, trained with a higher LoRA rank and a more detailed prompt, outperforms both ablated settings. These findings underscore the importance of prompt specificity and sufficient adaptation capacity during GRPO training.


\section{Keyframe Reward}
\label{sec:keyframe}
In this section, we propose and analyze a novel keyframe-based reward formulation designed for training under the GRPO objective. Prior work~\cite{egotextvqa} has emphasized the limitations of current vision-language models (VLMs) in accurately localizing keyframes that are crucial for answering questions in egocentric video tasks. To address this gap and promote temporal grounding, we introduce an additional reward signal that encourages the model to correctly identify salient or ``key" frames within the input video. This reward is used in conjunction with the previously defined accuracy and format rewards. An example of the training prompt used for this objective is provided in \autoref{tab:prompts}, where the model is instructed to list the keyframes within \textless frames\textgreater\ and \textless/frames\textgreater\ tags. \\
Due to the absence of ground-truth keyframe annotations in our training data, we adopt two proxy methods to estimate frame importance for each video-QA pair. Specifically, we treat each video as a sequence of frames $F_i$ where $i\in[0,n]$ and $n$ is the total number of frames in the video. The details of these proxy scoring methods are described below: \\
\begin{enumerate}
    \item \textbf{CLIP-score Approach}: The importance score $I(\cdot)$ for a video frame $F_i$, question $q$ and answer $a$ is given by \[I(F_i\vert q, a) = \text{cos\_sim}(\texttt{CLIP}_{\texttt{text}}(q+a), \texttt{CLIP}_{\texttt{image}}(F_i))\]
    where cos\_sim$()$ is the cosine similarity function, \texttt{CLIP}$_{\texttt{text}}$ is the CLIP~\cite{clip} text-encoder and \texttt{CLIP}$_{\texttt{image}}$ is the CLIP image-encoder. $+$ is a text append operator. 
    \item \textbf{Probability-score Approach}: The importance score $I(\cdot)$ for a video frame $F_i$, question $q$ and answer $a$ is given by \[I(F_i\vert q, a) = P_\theta(a \vert q, F_i)\] 
    where $P_\theta$ is the Qwen2.5-VL-7B-Instruct model. 
\end{enumerate}
The importance scores derived from the above methods are first normalized using a  $\texttt{Softmax}()$ function to yield a distribution over frames that reflects their relative relevance to the given question. These normalized scores are then used as a proxy for identifying ground-truth keyframes, which serve as supervision signals in our keyframe-based reward formulation. We explore three strategies for selecting these keyframes:
\begin{itemize}
    \item \textbf{\texttt{Top-K}}: Select $K$ frames with the highest importance scores. The corresponding training prompt is shown in \autoref{tab:prompts}, Row 3.
    \item \textbf{\texttt{Top-P}}: Select top $n$ frames with the highest importance scores whose cumulative score reaches atleast $P$, ensuring that only the most salient frames are retained.
    \item \textbf{\texttt{Bottom-P}}: Select $n$ frames with the lowest scores whose cumulative score reaches atleast $P$. Note that this strategy targets the identification of irrelevant frames rather than relevant ones. The corresponding training prompt is shown in \autoref{tab:prompts}, Row 4. We adopt this approach based on the observation that $\sim80\%$ of the frames fall within the Top-P = 0.8 threshold using both CLIP-based and probability-based scoring methods. This indicates that most frames are deemed important, making it potentially more tractable for the model to identify and discard irrelevant content.  
\end{itemize}

Finally, we define the keyframe reward as the F1-score between the predicted keyframes and the proxy ground-truth keyframes. This formulation encourages the model to achieve both high precision and high recall, discouraging trivial solutions such as predicting all input frames as keyframes. The results of this experiment are summarized in \autoref{tab:keyframe_results}. We observe that models trained with the keyframe reward exhibit strong performance across both egocentric benchmarks, even surpassing EgoVLM’s accuracy on the EgoPlan dataset. \\
However, a closer analysis of the model’s outputs reveals a concerning pattern: the model frequently predicts the same set of keyframe indices regardless of the input video. This behavior suggests that the model is not truly attending to the temporal content of the video and is failing to perform meaningful keyframe localization. We hypothesize that this arises from the model’s difficulty in distinguishing information between consecutive frames after they have been encoded by the vision encoder. This issue is further supported by the model's poor performance on simple diagnostic questions such as “How many frames are present in the provided video?”, which implies a lack of frame-level temporal awareness. 

\paragraph{Multi-image Input Format}
To address this limitation, we introduce an alternative input format in which the video is presented as a sequence of independent frame images. Each frame is prepended with a \textless frame\_$i$\textgreater\ tag where $i\in[1,n]$ denotes the index of a frame within a video of $n$ frames. We retrain the model using this multi-image input representation and evaluate its performance, as reported in \autoref{tab:multi_image_keyframe_results}. We see that by simply switching to this multi-image input format the base model now achieves a significant performance improvement on the egocentric benchmarks. Notably, it outperforms EgoVLM on the EgoPlan dataset, highlighting the importance of frame-level distinction in input representations for effective egocentric video understanding on these benchmarks. \\
However, incorporating keyframe rewards into the GRPO objective under this new input format leads to a degradation in performance on the EgoPlan and MVBench benchmarks. Moreover, the performance on EgoSchema lags significantly behind EgoVLM. This may be attributed to the increased complexity of the learning signal, where the model is now simultaneously optimizing for keyframe localization and task-specific accuracy. Additionally, the proxy nature of the keyframe supervision which is derived from unsupervised frame scoring methods rather than ground-truth annotations, may introduce noise into the reward signal, hindering effective policy updates during training. This highlights the need for more reliable keyframe annotations or improved methods for generating frame importance signals when incorporating temporal grounding objectives.

\begin{table}[ht!]
\centering
\small
\addtolength{\tabcolsep}{-5.pt}
\begin{tabular}{lccccc}
\toprule
\textbf{Model} & \textbf{EgoSchema}$\uparrow$ & \textbf{EgoPlan}$\uparrow$ & \textbf{MVBench}$\uparrow$ & \textbf{Avg.}$\uparrow$ \\ \midrule \midrule
Qwen2.5-VL &  59.4 & 32.9 & \textbf{66.9} & 53.1 \\ 
EgoVLM & \textbf{73.7} & 33.3 & 65.9 & \textbf{57.7} \\ 
CLIP \texttt{Top-K}=5 & 73.4 & 33.2 & 64.5 & 57.0 \\
Prob. \texttt{Top-K}=3 & 72.8 & 34.3 & 63.7 & 57.0 \\ 
Prob. \texttt{Top-K}=5 & 73.5 & \textbf{34.3} & 64.2 & 57.3 \\ 
Prob. \texttt{Top-P}=.5 & 72.5 & 32.8 & 63.8 & 56.4 \\ 
Prob. \texttt{Bot-P}=.2 & 73.3 & 33.2 & 65.0 & 57.2 \\ \bottomrule
\end{tabular}
\caption{Results (Accuracy in \%) of our keyframe reward experiments with video input format. CLIP and Prob. indicate CLIP and Probability-based importance scoring approaches respectively. \texttt{Bot-P} denotes the \texttt{Bottom-P} keyframe approach.}\looseness=-1
\label{tab:keyframe_results}
\end{table}

\begin{table}[ht!]
\centering
\small
\addtolength{\tabcolsep}{-5.pt}
\begin{tabular}{lccccc}
\toprule
\textbf{Model} & \textbf{EgoSchema}$\uparrow$ & \textbf{EgoPlan}$\uparrow$ & \textbf{MVBench}$\uparrow$ & \textbf{Avg.}$\uparrow$ \\ \midrule \midrule
\textit{Video Input} \\
Qwen2.5-VL &  59.4 & 32.9 & \textbf{66.9} & 53.1 \\
EgoVLM & \textbf{73.7} & 33.3 & 65.9 & \textbf{57.7} \\
\textit{Multi-image Input} \\
Qwen2.5-VL &  62.4 & \textbf{38.7} & 64.7 & 55.3 \\ 
Prob. \texttt{Top-K}=5 & 65.1 & 37.4 & 59.9 & 54.1 \\ 
Prob. \texttt{Top-P}=.5 & 64.8 & 36.8 & 59.4 & 53.7 \\ 
Prob. \texttt{Bot-P}=.2 & 65.3 & 37.0 & 61.1 & 54.5 \\ \bottomrule
\end{tabular}
\caption{Results (Accuracy in \%) of our keyframe reward experiments with multi-image and video input formats. CLIP and Prob. indicate CLIP and Probability-based importance scoring approaches respectively. \texttt{Bot-P} denotes the \texttt{Bottom-P} keyframe approach.}\looseness=-1
\label{tab:multi_image_keyframe_results}
\end{table}
\section{Conclusion and Future Work}
\label{sec:conclusion}
In this work, we present EgoVLM, a reasoning-centric vision-language model specifically designed for egocentric video understanding. By leveraging the GRPO training objective, we demonstrate that EgoVLM can effectively bridge the gap between visual perception and complex reasoning, achieving outstanding performance on multiple egocentric benchmarks. Unlike conventional SFT-based methods, our approach encourages deeper reasoning capabilities without relying on any chain-of-thought data, showcasing the efficacy of GRPO in first-person video domains. Furthermore, we show that EgoVLM’s reasoning ability generalizes well to non-egocentric settings, highlighting its broad applicability. Our ablation studies confirm the importance of prompt design and model capacity in successful GRPO training, and our exploration of keyframe-based rewards suggest future avenues for improving temporal grounding. \\
While EgoVLM demonstrates strong reasoning capabilities and competitive performance, several promising directions remain for future research. First, augmenting GRPO training with a curated chain-of-thought dataset tailored to egocentric contexts could further enhance the depth of the model’s reasoning. Additionally, integrating learnable visual encoders rather than frozen backbones may improve fine-grained understanding of occluded or subtle visual cues common in egocentric videos. Finally, our keyframe reward experiments suggest that more advanced frame-level grounding supervision, potentially derived from human annotations or attention mechanisms, could improve the model’s temporal grounding. 


{
    \small
    \bibliographystyle{ieeenat_fullname}
    \bibliography{main}
}


\end{document}